\documentclass[a4paper,11pt]{article}
\pdfoutput=1

\usepackage[a4paper, margin=1in]{geometry}
\usepackage{amsmath}
\usepackage{color}
\usepackage[pdftex, plainpages = false, pdfpagelabels, 
                 bookmarks=false,
                 bookmarksopen = true,
                 bookmarksnumbered = true,
                 breaklinks = true,
                 linktocpage,
                 pagebackref,
                 colorlinks = true,  
                 linkcolor = blue,
                 urlcolor  = blue,
                 citecolor = red,
                 anchorcolor = green,
                 hyperindex = true,
                 hyperfigures
                 ]{hyperref}

\title{SharpSAT-TD in Model Counting Competitions 2021-2023\thanks{Work funded by Adacemy of Finland under grants
322869 and 328718.}}
\author{Tuukka Korhonen\thanks{Department of Informatics, University of Bergen, Norway (\texttt{tuukka.korhonen@uib.no}). Work done while at University of Helsinki.} \and Matti J\"arvisalo\thanks{HIIT, Department of Computer Science, University of Helsinki, Finland (\texttt{matti.jarvisalo@helsinki.fi}).}}

\date{}

\begin{document}

\maketitle

\begin{abstract}
\noindent
We describe SharpSAT-TD, our submission to the unweighted and weighted tracks of the Model Counting Competition\footnote{See \url{https://mccompetition.org/}.} in 2021-2023, which has won in total $6$ first places in different tracks of the competition.
SharpSAT-TD is based on SharpSAT~[Thurley,~SAT~2006], with the primary novel modification being the use of tree decompositions in the variable selection heuristic as introduced by the authors in~[CP~2021].
Unlike the version of SharpSAT-TD evaluated in~[CP~2021], the current version that is available in \url{https://github.com/Laakeri/sharpsat-td} features also other significant modifications compared to the original SharpSAT, for example, a new preprocessor.
\end{abstract}

\section{Overview}
SharpSAT-TD is based on the exact model counter SharpSAT~\cite{DBLP:conf/sat/Thurley06}, from which it inherits the basic structure of a search-based model counter with clause learning, component analysis, and component caching.
The main new feature in SharpSAT-TD is that we compute a tree decomposition of the input formula with the FlowCutter algorithm~\cite{DBLP:journals/jea/HamannS18,DBLP:journals/corr/abs-1709-08949}, and integrate the tree decomposition to the variable selection heuristic of the counter by a method introduced by the authors in~\cite{DBLP:conf/cp/KorhonenJ21}.
Another significant new feature is a new preprocessor.
Further, SharpSAT-TD extends SharpSAT by directly supporting weighted model counting.

SharpSAT-TD is available in \url{https://github.com/Laakeri/sharpsat-td}.
We note that the current version of SharpSAT-TD differs significantly from the version evaluated in~\cite{DBLP:conf/cp/KorhonenJ21}, as the version evaluated in~\cite{DBLP:conf/cp/KorhonenJ21} differed from SharpSAT only in the variable selection heuristic, while the current SharpSAT-TD has also other new features.
The current SharpSAT-TD has stayed almost unchanged since the Model Counting Competition 2021: After the competition we fixed some bugs and added proper support for weighted model counting, but after that no further updates have been made.

\section{Integrating Tree Decompositions into the Variable Selection Heuristic}
In this section we briefly outline the method from~\cite{DBLP:conf/cp/KorhonenJ21} as used in SharpSAT-TD.

We compute a tree decomposition of the primal graph of the input formula with the anytime FlowCutter algorithm~\cite{DBLP:journals/jea/HamannS18,DBLP:journals/corr/abs-1709-08949}, always using 120 seconds in the computation.
To make the tree decomposition rooted, we choose as the root a node so that the bag of the root is a balanced separator of the primal graph.

In SharpSAT, the variable selection heuristic is based on frequency and activity scores.
In particular, for each variable $x$ the scores $\texttt{freq}(x)$ and $\texttt{act}(x)$ are maintained, and the variable with the highest $\texttt{score}(x) = \texttt{freq}(x) + \texttt{act}(x)$ is selected.
We add a tree decomposition based term to this score.
In SharpSAT-TD the variable selection is done using 
\[\texttt{score}(x) = \texttt{freq}(x) + \texttt{act}(x) - C \cdot d(x),\]
where $d(x)$ denotes the distance in the tree decomposition from the root node to a closest node whose bag contains $x$ (normalized to the interval $[0 \ldots 1]$), and $C$ is some positive constant.
The variable $x$ with the highest $\texttt{score}(x)$ is selected first for branching.
In particular, this score prefers variables that are closer to the root in the tree decomposition.
The constant $C$ is selected as $C = 100 \exp(n/w)/n$, where $w$ is the width of the tree decomposition and $n$ is the number of variables.
Note that this makes the tree decomposition based score more significant when the width $w$ is small compared to $n$.

\section{Preprocessing}
We implement a new preprocessor completely from scratch.
The preprocessing is done before computing the tree decomposition, and one of the goals of the preprocessing is to decrease the treewidth of the formula.
The preprocessing techniques used, in the order of application, are propagation-based vivification, complete vivification, sparsification, equivalence merging, and our own implementation of B+E~\cite{DBLP:conf/ijcai/LagniezLM16,DBLP:conf/ecai/PietteHS08}.

In propagation-based vivification, for each clause $c$ and a literal $l$ in $c$ we check if $\lnot (c \setminus \{l\})$ implies UNSAT via unit propagation, and if yes we strenghten $c$ by removing $l$ from it.
In complete vivification we do the same, but the UNSAT check is done by a complete SAT-solver.
The SAT-solver is a new implementation of a CDCL-based SAT-solver, with techniques in~\cite{DBLP:conf/ijcai/AudemardS09,sorensson2005minisat} implemented, and specifically optimized for the setting of model counting preprocessing.
Note that complete vivification also results in backboning the formula.
In sparsification we attempt to remove clauses that are implied by other clauses.
The redundancy of a clause is checked with a SAT-solver.
The goal of sparsification is to reduce treewidth.
In equivalence merging we merge two variables if they are equivalent and they are adjacent in the primal graph.
The equivalency is checked with a SAT-solver.
Note that in the primal graph, merging two adjacent variables corresponds to edge contraction, which does not increase treewidth.
Finally, for unweighted formulas we re-implement the B+E algorithm~\cite{DBLP:conf/ijcai/LagniezLM16}.
Our implementation of B+E is done in a way to ensure to not increase the treewidth of the formula.
On some instances our implemention of B+E seems faster than the original, while eliminating the same number of variables.

\section{Further New Features and Modifications}
We disabled the ``implicit BCP'' feature of SharpSAT because it decreased the overall number of public instances solved in preliminary experiments, although we note that on some instances it appears useful.

We changed the SharpSAT learned clause scoring scheme into the LBD scheme~\cite{DBLP:conf/ijcai/AudemardS09}.
This also fixed a bug of SharpSAT where it does not delete any learned clauses if their median score is 0, which often happened.
SharpSAT-TD also changes the desired number of learned clauses stored based on their estimated usefullness, i.e., LBD scores.
In preliminary experiments it seems that these LBD score based techniques do not result in very significant effects in the running times, but the bugfix sometimes does.

We implemented the probabilistic component caching scheme introduced in GANAK~\cite{DBLP:conf/ijcai/SharmaRSM19}.
We note that even though GANAK is also a SharpSAT-derivant, our implementation is different from the implementation of GANAK.
Also, in our implementation instead of having a dynamic hash-length we fix a hash-length of 128 bits, noting that it results in a collision probability of $< 10^{-9}$ on all cases where the counter does at most $10^{14}$ cache lookups (note that doing more than $10^{14}$ cache lookups within the 3600 seconds time limit of the competition seems impossible on the competition hardware).

\section{Extension to Weighted Model Counting}
While in principle it is clear that any model counter whose trace corresponds to a d-DNNF-compilation can be extended to also weighted model counting, efficient implementation is not necessarily straightforward, and in the case of SharpSAT required editing hundreds of lines of code.
Our extension to weighted model counting is implemented via template parameters, so changing the data types or the semiring that the counter works on is simple.

\bibliography{desc}
\bibliographystyle{plain}

\end{document}